\documentclass{article} 
\usepackage{iclr2017_conference,times}
\usepackage{hyperref}
\usepackage{url}
\usepackage{graphicx}
\usepackage{epstopdf}
\usepackage{fouridx} 
\usepackage{amsmath}

\title{Deep Multi-task Representation Learning: \\A Tensor Factorisation Approach}

\author{Yongxin Yang, Timothy M. Hospedales\\
Queen Mary, University of London\\
\texttt{\{yongxin.yang, t.hospedales\}@qmul.ac.uk} \\
}

\iclrfinalcopy 

\begin{document}

\maketitle

\begin{abstract}
Most contemporary multi-task learning methods assume linear models. This setting is considered \emph{shallow} in the era of deep learning. In this paper, we present a new deep multi-task representation learning framework that learns cross-task sharing structure \emph{at every layer in a deep network}. Our approach is based on generalising the matrix factorisation techniques explicitly or implicitly used by many conventional MTL algorithms to tensor factorisation, to realise automatic learning of end-to-end knowledge sharing in deep networks. This is in contrast to existing deep learning approaches that need a user-defined multi-task sharing strategy. Our approach applies to both homogeneous and heterogeneous MTL. Experiments demonstrate the efficacy of our deep multi-task representation learning in terms of both higher accuracy and fewer design choices.
\end{abstract}

\section{Introduction}

The paradigm of multi-task learning is to learn multiple related tasks simultaneously so that knowledge obtained from each task can be re-used by the others. Early work in this area focused on neural network models \citep{Caruana1997}, while more recent methods have shifted focus to kernel methods, sparsity and low-dimensional task representations of linear models \citep{Evgeniou2004, Argyriou2008, daume2012gomtl}. Nevertheless given the impressive practical efficacy of contemporary deep neural networks (DNN)s in many important applications, we are motivated to revisit MTL from a deep learning perspective. 

While the machine learning community has focused on MTL for shallow linear models recently, applications have continued to exploit neural network MTL \citep{ZhangLLT14, Liu15nlp}. The typical design pattern dates back at least 20 years \citep{Caruana1997}: define a DNN with shared lower representation layers, which then forks into separate layers and losses for each task. The sharing structure is defined manually: full-sharing up to the fork, and full separation after the fork. However this complicates DNN architecture design because the user must specify the sharing structure: How many task specific layers? How many task independent layers? How to structure sharing if there are many tasks of varying relatedness?

In this paper we present a method for end-to-end multi-task learning in DNNs. This contribution can be seen as generalising shallow MTL methods \citep{Evgeniou2004,Argyriou2008,daume2012gomtl} to learning how to share \emph{at every layer} of a deep network; or  as learning the sharing structure for deep MTL \citep{Caruana1997,ZhangLLT14,factortensorrnn14,Liu15nlp} which currently must be defined manually on a problem-by-problem basis.

Before proceeding it is worth explicitly distinguishing some different problem settings, which have all been loosely referred to as MTL in the literature. \textbf{Homogeneous MTL:} Each task corresponds to a \emph{single} output. For example, MNIST digit recognition is commonly used to evaluate MTL algorithms by casting it as 10 binary classification tasks \citep{daume2012gomtl}. \textbf{Heterogeneous MTL:} Each task corresponds to a unique set of output(s) \citep{ZhangLLT14}. For example, one may want simultaneously predict a person's age (task one: multi-class classification or regression) as well as identify their gender (task two: binary classification) from a face image. 

In this paper, we propose a multi-task learning method that works on all these settings. The key idea is to use tensor factorisation to divide each set of model parameters (i.e., both FC weight matrices, and convolutional kernel tensors) into \emph{shared} and \emph{task-specific} parts. It is a natural generalisation of shallow MTL methods that explicitly or implicitly are based on matrix factorisation \citep{Evgeniou2004, Argyriou2008, daume2012gomtl,iii2007frustratingly}. As linear methods, these typically require pre-engineered features. In contrast, as a deep network, our generalisation can learn directly from raw image data, determining sharing structure in a layer-wise fashion. For the simplest NN architecture -- no hidden layer, single output -- our method reduces to matrix-based ones, therefore matrix-based methods including \citep{Evgeniou2004, Argyriou2008, daume2012gomtl, iii2007frustratingly} are special cases of ours.

\section{Related Work}

\textbf{Multi-Task Learning}\quad 
Most contemporary MTL algorithms assume that the input and model are both $D$-dimensional vectors. The models of  $T$ tasks can then be stacked into a $D\times T$ sized matrix $W$. Despite different motivations and implementations, many matrix-based MTL methods work by placing constrains on $W$. For example,  posing an $\ell_{2,1}$ norm on $W$ to encourage low-rank $W$ \citep{Argyriou2008}. Similarly, \citep{daume2012gomtl} factorises $W$ as $W=LS$, i.e., it  assigns a lower rank as a hyper-parameter. An earlier work \citep{Evgeniou2004} proposes that the linear model for each task $t$ can be written as $w_t = \hat{w}_t + \hat{w}_0$. This is the factorisation $L = [\hat{w}_0, \hat{w}_1, \dots, \hat{w}_T]$ and $S = [\mathbf{1}_{1\times T}; \mathbf{I}_T]$. In fact, such matrix factorisation encompasses many MTL methods. E.g., \citep{Xue2007}  assumes $S_{\cdot, i}$ (the $i$th column of $S$) is a unit vector generated by a Dirichlet Process and \citep{daume2012flexiblemtl} models $W$ using linear factor analysis with Indian Buffet Process \citep{Griffiths2011IBP} prior on $S$. 

\textbf{Tensor  Factorisation}\quad In deep learning, tensor factorisation has been used to exploit factorised tensors' fewer parameters than the original (e.g., 4-way convolutional kernel) tensor, and thus compress and/or speed up the model, e.g., \citep{LebedevGROL14, novikov15tensornet}. For shallow linear MTL, tensor factorisation has been used to address problems where tasks are described by multiple independent factors rather than merely indexed by a single factor \citep{yang15}. Here the $D$-dimensional linear models for all unique tasks stack into a tensor $\mathcal{W}$, of e.g. $D\times T_1 \times T_2$  in the case of two task factors. Knowledge sharing is then achieved by imposing tensor norms on $\mathcal{W}$ \citep{icml2013_romera-paredes13,wimalawarnemultitask}. Our framework factors tensors for the different reason that for DNN models,  parameters include  convolutional kernels ($N$-way tensors) or $D_1\times D_2$ FC layer weight matrices ($2$-way tensors). Stacking up these parameters for many tasks results in $D_1\times\dots\times D_N\times T$ tensors within which we share knowledge through factorisation.

\textbf{Heterogeneous MTL and DNNs}\quad 
Some studies consider heterogeneous MTL, where tasks may have different numbers of outputs \citep{Caruana1997}. This differs from the previously discussed studies \citep{Evgeniou2004, Argyriou2008, bonilla2007multi, jacob2009clustered, daume2012gomtl, icml2013_romera-paredes13,wimalawarnemultitask} which implicitly assume that each task has a single output. Heterogeneous MTL typically uses neural networks with multiple sets of outputs and losses. E.g., \cite{Huang2013cross} proposes a shared-hidden-layer DNN model for multilingual speech processing, where each task corresponds to an individual language. \cite{ZhangLLT14} uses a DNN to find facial landmarks (regression) as well as recognise facial attributes (classification); while \cite{Liu15nlp} proposes a DNN for query classification and information retrieval (ranking for web search). A key commonality of these studies is that they all require a user-defined parameter sharing strategy. A typical design pattern is to use shared layers (same parameters) for lower layers of the DNN and then split (independent parameters) for the top layers. However, there is no systematic way to make such design choices, so researchers usually rely on trial-and-error, further complicating the already somewhat dark art of DNN design. In contrast, our method learns where and how much to share representation parameters across the tasks, hence significantly reducing the space of DNN design choices.

\textbf{Parametrised DNNs}\quad
Our MTL approach is a \emph{parameterised DNN} \citep{sigaud2015gatedInventory}, in that DNN weights are dynamically generated given some side information  -- in the case of MTL, given the task identity. In a related example of speaker-adaptive speech recognition \citep{Tan2016cluster} there may be several clusters in the data (e.g., gender, acoustic conditions), and each speaker's model could be a linear combination of these latent task/clusters' models. They model each speaker $i$'s weight matrix $W^{(i)}$ as a sum of $K$ base models $\tilde{W}$, i.e., $W^{(i)}=\sum_{k=1}^{K} \lambda_p^{(i)} \tilde{W}^{(p)}$. The difference between speakers/tasks comes from $\lambda$ and the base models are shared. An advantage of this is that, when new data come, one can choose to re-train $\lambda$ parameters only, and keep $\tilde{W}$ fixed. This will significantly reduce the number of parameters to learn, and consequently the required training data. Beyond this, \cite{yang15} show that it is possible to train another neural network to \emph{predict} those $\lambda$ values from some abstract metadata. Thus a model for an \emph{unseen} task can be generated on-the-fly with \emph{no} training instances given an abstract description of the task. The techniques developed here are compatible with both these ideas of generating models with minimal or no effort.

\section{Methodology}

\subsection{Preliminaries}

We first recap some tensor factorisation basics before explaining how to factorise DNN weight tensors for multi-task representation learning. An $N$-way tensor $\mathcal{W}$ with shape $D_1 \times D_2 \times \cdots D_N$ is an $N$-dimensional array containing $\prod_{n=1}^{N} D_n$ elements. Scalars, vectors, and matrices can be seen as $0$, $1$, and $2$-way tensors respectively, although the term tensor is usually used for $3$-way or higher. 
A mode-$n$ fibre of $\mathcal{W}$ is a $D_n$-dimensional vector obtained by fixing all but the $n$th index. The mode-$n$ flattening $W_{(n)}$ of $\mathcal{W}$ is the matrix of size $D_n \times \prod_{i\lnot n} D_i$ constructed by concatenating all of the $\prod_{i\lnot n} D_i$ mode-$n$ fibres along columns. 

The dot product of two tensors is a natural extension of matrix dot product, e.g., if we have a tensor $\mathcal{A}$ of size $M_1 \times M_2 \times \cdots P$ and a tensor $\mathcal{B}$ of size $P \times N_1 \times N_2 \dots$, the tensor dot product $\mathcal{A} \bullet \mathcal{B}$ will be a tensor of size $M_1 \times M_2 \times \cdots N_1 \times N_2 \cdots$ by matrix dot product $A_{(-1)}^T B_{(1)}$ and reshaping\footnote{We slightly abuse `-1' referring to the last axis of the tensor.}. More generally, tensor dot product can be performed along specified axes, $\mathcal{A}\bullet\fourIdx{}{(i,j)}{}{}{\mathcal{B}} = A_{(i)}^T B_{(j)}$ and reshaping. Here the subscripts indicate the axes of $\mathcal{A}$ and $\mathcal{B}$ at which dot product is performed. E.g., when $\mathcal{A}$ is of size $M_1 \times P \times M_3 \times \cdots M_I$ and $\mathcal{B}$ is of size $N_1 \times N_2 \times P \times \cdots N_J$, then  $\mathcal{A}\bullet\fourIdx{}{(2,3)}{}{}{\mathcal{B}}$ is a tensor of size $M_1 \times M_3 \times \cdots M_I \times N_1 \times N_2 \times \cdots N_J$. 

\vspace{0.1cm}\noindent\textbf{Matrix-based Knowledge Sharing}\quad Assume we have $T$ linear models (tasks) parametrised by $D$-dimensional weight vectors, so the collection of all models forms a size $D\times T$ matrix $W$.
One commonly used MTL approach \citep{daume2012gomtl} is to place a structure constraint on $W$, e.g., $W = LS$, where $L$ is a $D\times K$ matrix and $S$ is a $K\times T$ matrix. This factorisation recovers a \emph{shared} factor $L$ and a \emph{task-specific} factor $S$. One can see the columns of $L$ as latent basis tasks, and the model  $w^{(i)}$ for the $i$th task is the linear combination of those latent basis tasks with task-specific information $S_{\cdot,i}$.
\begin{equation}
\label{eq:gomtl}
w^{(i)} := W_{\cdot, i} = LS_{\cdot, i} = \sum_{k=1}^{K} L_{\cdot, k} S_{k,i}
\end{equation}
\vspace{0.1cm}\noindent\textbf{From Single to Multiple Outputs}\quad 
Consider extending this matrix factorisation approach to the case of multiple outputs. The model for each task is then a $D_1 \times D_2$ matrix, for $D_1$ input and $D_2$ output dimensions. The collection of all those matrices constructs a $D_1 \times D_2 \times T$ tensor. A straightforward extension of Eq.~\ref{eq:gomtl} to this case is
\begin{equation}
\label{eq:laf_mtl}
W^{(i)} := \mathcal{W}_{\cdot, \cdot, i} = \sum_{k=1}^{K} \mathcal{L}_{\cdot,\cdot, k} S_{k,i}
\end{equation}
This is equivalent to imposing the same structural constraint on $W_{(3)}^T$ (transposed mode-$3$ flattening of $\mathcal{W}$). It is important to note that this allows knowledge sharing across the tasks \emph{only}. I.e., knowledge sharing is only across-tasks not across dimensions within a task. However it may be that the knowledge learned in the mapping to one output dimension may be useful to the others within one task. E.g., consider recognising photos of handwritten and print digits -- it may be useful to share across handwritten-print; as well as across different digits within each. In order to support general knowledge sharing across both tasks and outputs within tasks, we propose to use more general tensor factorisation techniques. Unlike for matrices, there are multiple definitions of tensor factorisation, and we use Tucker \citep{Tuck1966c} and Tensor Train (TT) \citep{Oseledets2011} decompositions.

\subsection{Tensor Factorisation for Knowledge Sharing}

\vspace{0.1cm}\noindent\textbf{Tucker Decomposition}\quad Given an $N$-way tensor of size $D_1\times D_2 \dots \times D_N$, Tucker decomposition outputs a core tensor $\mathcal{S}$ of size $K_1\times K_2 \dots \times K_N$, and $N$ matrices $U^{(n)}$ of size $D_n \times K_n$, such that,

\begin{eqnarray}
\mathcal{W}_{d_1,d_2,\dots,d_N} & = & \sum_{k_1=1}^{K_1}\sum_{k_2=1}^{K_2}\cdots\sum_{k_N=1}^{K_N} \mathcal{S}_{k_1,k_2,\dots,k_N}{U}^{(1)}_{d_1,k_1}{U}^{(2)}_{d_2,k_2}\cdots {U}^{(N)}_{d_N,k_N}\label{eq:tucker}\\
\mathcal{W} & = & \mathcal{S} \bullet {\fourIdx{}{(1,2)}{(1)}{}{U}} \bullet {\fourIdx{}{(1,2)}{(2)}{}{U}} \cdots \bullet {\fourIdx{}{(1,2)}{(N)}{}{U}}\label{eq:tucker2}
\end{eqnarray}

Tucker decomposition is usually implemented by an alternating least squares (ALS) method \citep{Kolda2009}. However  \citep{Lathauwer2000} treat it as a higher-order singular value decomposition (HOSVD), which is more efficient to solve: $U^{(n)}$ is exactly the $U$ matrix from the SVD of mode-$n$ flattening $W_{(n)}$ of $\mathcal{W}$, and the core tensor $\mathcal{S}$ is obtained by,

\begin{equation}
\label{eq:hosvd}
\mathcal{S} = \mathcal{W} \bullet \fourIdx{}{(1,1)}{(1)}{}{U} \bullet \fourIdx{}{(1,1)}{(2)}{}{U} \cdots \bullet \fourIdx{}{(1,1)}{(N)}{}{U}
\end{equation}

\vspace{0.1cm}\noindent\textbf{Tensor Train Decomposition}\quad Tensor Train (TT) Decomposition outputs $2$ matrices $U^{(1)}$ and $U^{(N)}$ of size $D_1\times K_1$ and $K_{N-1} \times D_N$ respectively, and $(N-2)$ $3$-way tensors $\mathcal{U}^{(n)}$ of size $K_{n-1} \times D_n \times K_n$. The elements of $\mathcal{W}$ can be computed by,

\begin{eqnarray}
\mathcal{W}_{d_1,d_2,\dots,d_N} &=& \sum_{k_1=1}^{K_1}\sum_{k_2=1}^{K_2}\cdots\sum_{k_{N-1}=1}^{K_{N-1}}{U}^{(1)}_{d_1,k_1}\mathcal{U}^{(2)}_{k_1,d_2,k_2}\mathcal{U}^{(3)}_{k_2,d_3,k_3}\cdots{U}^{(N)}_{k_{N-1},d_N}\label{eq:tt}\\
&=& {U}^{(1)}_{d_1,\cdot}\mathcal{U}^{(2)}_{\cdot,d_2,\cdot}\mathcal{U}^{(3)}_{\cdot,d_3,\cdot}\cdots{U}^{(d)}_{\cdot,d_N}\label{eq:tt2}\\
\mathcal{W} &=& {U}^{(1)} \bullet \mathcal{U}^{(2)} \cdots \bullet {U}^{(N)}\label{eq:tt3}
\end{eqnarray}

\noindent where $\mathcal{U}^{(n)}_{\cdot,d_n,\cdot}$ is a matrix of size $K_{n-1} \times K_n$ sliced from $\mathcal{U}^{(n)}$ with the second axis fixed at $d_n$. The TT decomposition is typically realised with a recursive SVD-based solution \citep{Oseledets2011}.

\textbf{Knowledge Sharing}\quad If the final axis of the input tensor above indexes tasks, i.e. if $D_N=T$  then the last factor $U^{(N)}$ in both decompositions encodes a matrix of task specific knowledge, and the other factors encode shared knowledge.

\subsection{Deep Multi-Task Representation Learning}

To realise deep multi-task representation learning (DMTRL), we learn one DNN per-task each with the same architecture\footnote{Except  heterogeneous MTL, where the output layer is necessarily unshared due to different dimensionality.}. However each corresponding layer's weights are generated with one of the knowledge sharing structures in Eq.~\ref{eq:laf_mtl}, Eq.~\ref{eq:tucker2} or Eq.~\ref{eq:tt3}. It is important to note that we apply these `right-to-left' in order to generate weight tensors with the specified sharing structure, rather than actually applying Tucker or TT to decompose an input tensor. In the forward pass, we synthesise weight tensors $\mathcal{W}$ and perform inference as usual, so the method can be thought of as tensor \emph{composition} rather than decomposition.

Our weight generation (construct tensors from smaller pieces) does not introduce non-differentiable terms, so our deep multi-task representation learner is trainable via standard backpropagation.
Specifically, in the backward pass over FC layers, rather than directly learning the 3-way tensor $\mathcal{W}$, our methods learn either $\{\mathcal{S}, U_1, U_2, U_3\}$ (DMTRL-Tucker, Eq.~\ref{eq:tucker2}),  $\{U_1, \mathcal{U}_2, U_3\}$ (DMTRL-TT, Eq.~\ref{eq:tt3}), or in the simplest case $\{\mathcal{L},S\}$ (DMTRL-LAF\footnote{LAF refers to Last Axis Flattening.}, Eq.~\ref{eq:laf_mtl}). Besides FC layers, contemporary DNN designs often exploit convolutional layers. Those layers usually contain kernel filter parameters that are $3$-way tensors of size $H\times W \times C$, (where $H$ is  height, $W$ is  width, and $C$ is the number of input channels) or $4$-way tensors of size $H\times W \times C \times M$, where $M$ is the number of filters in this layer (i.e., the number of output channels). The proposed methods naturally extend to convolution layers as convolution just adds more axes on the left-hand side. E.g., the collection of parameters from a given convolutional layer of $T$ neural networks forms a tensor of shape $H\times W \times C \times M \times T$.

\begin{figure}[t]
\centering
\includegraphics[width=0.9\linewidth]{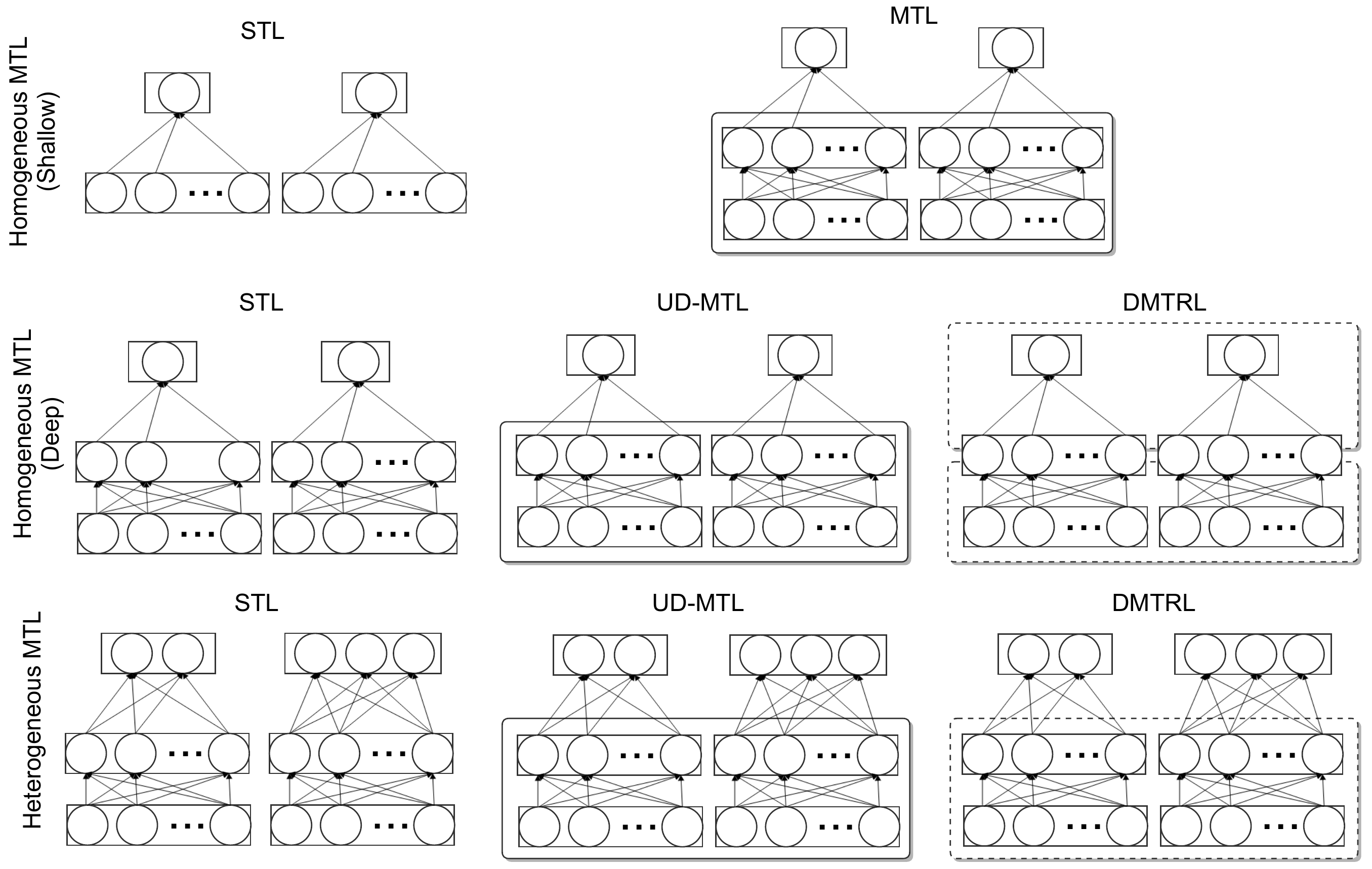}
{\small
\caption[Illustration]{Illustrative example with two tasks corresponding to two neural networks in homogeneous (single output) and heterogeneous (different output dimension) cases. Weight layers grouped by solid rectangles are tied across networks. Weight layers grouped by dashed rectangles are softly shared across networks with our method. Ungrouped weights are independent.\\
Homogeneous MTL Shallow: Left is STL (two independent networks); right is MTL. In the case of vector input and no hidden layer, our method is equivalent to conventional matrix-based MTL methods.
Homogeneous MTL Deep: STL (Left) is independent networks. User-defined-MTL (UD-MTL)  selects layers to share/separate. Our DMTRL learns sharing at every layer.
Heterogeneous MTL: UD-MTL selects layers to share/separate. DMTRL learns sharing at every shareable layer.
}\label{fig:demo}
}
\end{figure}

These knowledge sharing strategies provide a way to \emph{softly} share parameters across the corresponding layers of each task's DNN: where, what, and how much to share are learned from data. This is in contrast to the conventional Deep-MTL approach of manually selecting a set of layers to undergo \emph{hard} parameter sharing: by tying weights so each task uses exactly the same weight matrix/tensor for the corresponding layer \citep{ZhangLLT14, Liu15nlp}; and a set of layers to be completely separate: by using independent weight matrices/tensors. In contrast our approach benefits from: (i) automatically learning this sharing structure from data rather than requiring user trial and error, and (ii) smoothly interpolating between fully shared and fully segregated layers, rather than a hard switching between these states. An illustration of the proposed framework for different problem settings can be found in Fig.~\ref{fig:demo}.

\section{Experiments}

\textbf{Implementation Details}\quad Our method is implemented with TensorFlow \citep{tensorflow2015}. The code is released on GitHub\footnote{\url{https://github.com/wOOL/DMTRL}}. For DMTRL-Tucker, DMTRL-TT, and DMTRL-LAF, we need to assign the rank of each weight tensor. The DNN architecture itself may be complicated and so can benefit from different ranks at different layers, but grid-search is impractical. However, since both Tucker and TT decomposition methods have SVD-based solutions, and vanilla SVD is directly applicable to DMTRL-LAF, we can initialise the model and set the ranks as follows: First train the DNNs independently in single task learning mode. Then pack the layer-wise parameters as the input for tensor decomposition. When SVD is applied, set a threshold for relative error so SVD will pick the appropriate rank. Thus our method needs only a \emph{single} hyper parameter of max reconstruction error (we set to $\epsilon=10\%$ throughout) that indirectly specifies the ranks of every layer. Note that training from random initialisation also works, but the STL-based initialisation makes rank selection easy and transparent. Nevertheless, like \citep{daume2012gomtl} the framework is not sensitive to rank choice so long as they are big enough. If random initialisation is desired to eliminate the pre-training requirement, good practice is to initialise parameter tensors by a suitable random weight distribution first, then do decomposition, and use the decomposed values for initialising the factors (the \emph{real} learnable parameters in our framework). In this way, the resulting re-composed tensors will have approximately the intended distribution. Our sharing is applied to weight parameters only, bias terms are not shared. Apart from initialisation, decomposition is not used anywhere.

\subsection{Homogeneous MTL}
\textbf{Dataset, Settings and Baselines} We use MNIST handwritten digits. The task is to recognise digit images zero to nine. When this dataset is used for the evaluation of MTL methods, ten 1-vs-all binary classification problems usually define ten tasks \citep{daume2012gomtl}. The dataset has a given train (60,000 images) and test (10,000 images) split. Each instance is a monochrome image of size $28\times 28\times 1$.

We use a modified LeNet \citep{lecun98lenet} as the CNN architecture. The first convolutional layer has $32$ filters of size $5\times 5$, followed by $2\times 2$ max pooling. The second convolutional layer has $64$ filters of size $4\times 4$, and again a $2 \times 2$ max pooling. After these two convolutional layers, two fully connected layers with $512$ and $1$ output(s) are placed sequentially.  The convolutional  and  first FC layer use RELU $f(x)=\max(x,0)$ activation function. We use hinge loss, $\ell(y) = \max(0, 1-\hat{y} \cdot y)$, where  $y\in\pm 1$ is the true label and $\hat{y}$ is the output of each task's neural network.

Conventional matrix-based MTL methods \citep{Evgeniou2004, Argyriou2008, daume2012gomtl, icml2013_romera-paredes13,wimalawarnemultitask} are linear models taking vector input only, so they need a preprocessing that flattens the image into a vector, and typically reduce dimension by PCA. As per our motivation for studying \emph{Deep} MTL, our methods decisively outperform such shallow linear baselines. Thus to find a stronger MTL competitor, we instead search user defined architectures for Deep-MTL parameter sharing (cf \citep{ZhangLLT14, Liu15nlp,Caruana1997}).
In all of the four parametrised layers (pooling has no parameters), we set the first $N$ ($1\le N\le 3$) to be \emph{hard} shared\footnote{This is not strictly all possible user-defined sharing options. For example, another possibility is the first convolutional layer and the first FC layer could be fully shared, with the second convolutional layer being independent (task specific). However, this is against the intuition that lower/earlier layers are more task agnostic, and later layers more task specific. Note that sharing the last layer is technically possible but not intuitive, and in any case not meaningful unless at least one early layer is unshared, as the tasks are different.}. We then use cross-validation to select among the three user-defined MTL architectures and the best option is $N=3$, i.e., the first three layers are fully shared (we denote this model UD-MTL).
For our methods, all four parametrised layers are softly shared with the different factorisation approaches. To evaluate different MTL methods and a baseline of single task learning (STL), we take ten different fractions of the given 60K training split, train the model, and test on the 10K testing split. For each fraction, we repeat the experiment $5$ times with randomly sampled training data. We report two performance metrics: (1) the mean error rate of the ten binary classification problems and (2) the error rate of recognising a digit by ranking each task's 1-vs-all output (multi-class classification error).

\begin{figure}[t]
\centering
\includegraphics[width=0.85\linewidth]{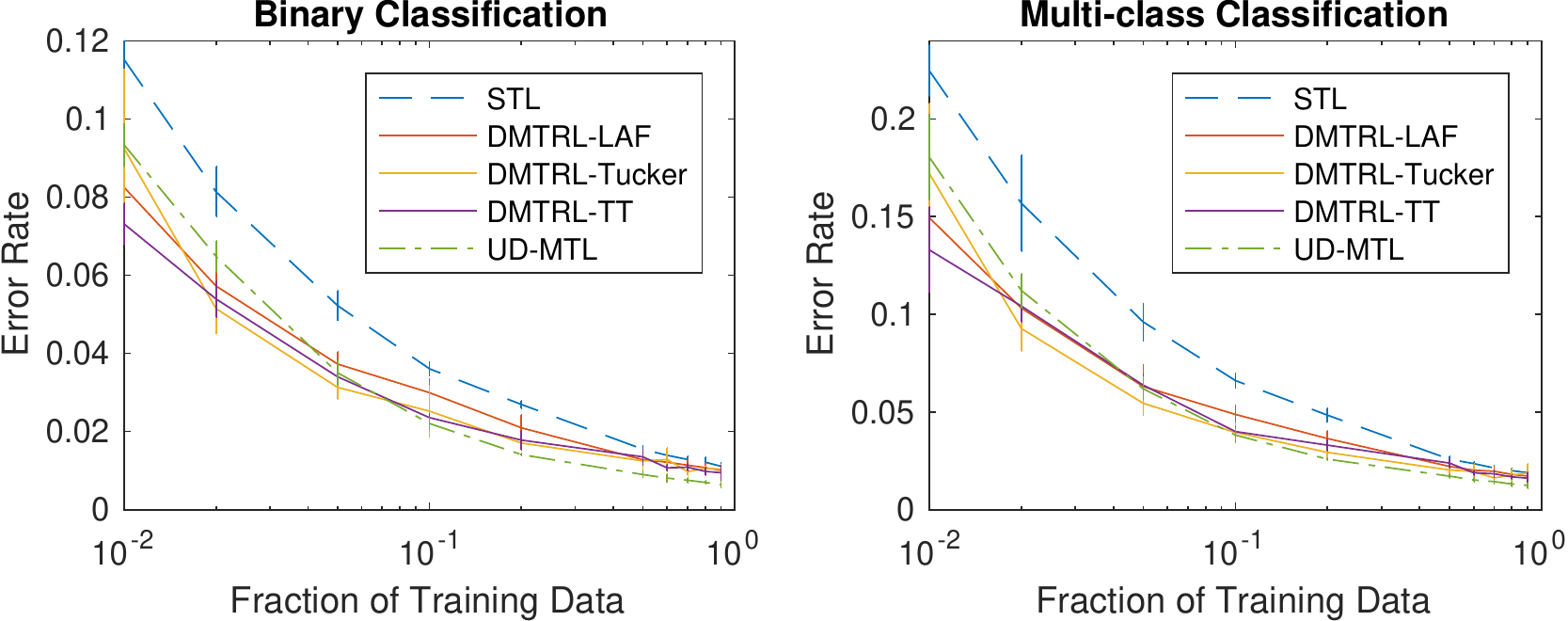}
\caption{Homogeneous MTL: digit recognition on MNIST dataset. Each digit provides a task.}
\label{fig:mnist}
\end{figure}

\textbf{Results}\quad
As we can see in Fig.~\ref{fig:mnist}, all MTL approaches outperform STL, and the advantage is more significant when the training data is small. The proposed methods, DMTRL-TT and DMTRL-Tucker outperform the best user-defined MTL when the training data is very small, and their performance is comparable when the training data is large. 

\textbf{Further Discussion}\quad
For a slightly unfair comparison, in the case of binary classification with 1000 training data, shallow matrix-based MTL methods with PCA feature \citep{Kang2011learning,daume2012gomtl} reported $14.0\%$ / $13.4\%$ error rate. With the same amount of data, our methods have error rate below $6\%$. This shows the importance of our deep end-to-end multi-task representation learning contribution versus conventional shallow MTL. Since the error rates in \citep{Kang2011learning, daume2012gomtl} were produced on a private subset of MNIST dataset with PCA representations only, to ensure a direct comparison, we implement several classic MTL methods and compare them in Appendix~\ref{appx2}.

For readers interested in the connection to model capacity (number of parameters), we present further analysis in Appendix~\ref{appx1}.

\subsection{Heterogeneous MTL: Face Analysis}

\textbf{Dataset, Settings and Baselines}\quad The AdienceFaces \citep{Eidinger14} is a large-scale face images dataset with the labels of each person's gender and age group. We use this dataset for the evaluation of heterogeneous MTL with two tasks: (i) gender classification (two classes) and (ii) age group classification (eight classes). Two independent CNN models for this benchmark are introduced in \citep{Levi15}. The two CNNs have the same architecture except for the last  fully-connected layer, since the heterogeneous tasks have different number of outputs (two / eight). We take  these CNNs from \citep{Levi15} as the STL baseline.
We again search for the best possible user-defined MTL architecture as a strong competitor: the proposed CNN has six layers -- three convolutional and three fully-connected layers. The last fully-connected layer has non-shareable parameters because they are of different size. To search the MTL design-space, we try setting the first $N$ ($1\leq N \leq 5$) layers to be hard shared between the tasks. Running 5-fold cross-validation on the train set to evaluate the architectures, we find the best choice is $N=5$ (i.e., all layers fully shared before the final heterogeneous outputs). For our proposed methods, all the layers before the last heterogeneous dimensionality FC layers are softly shared.

We select increasing fractions of the AdienceFaces train split randomly, train the model, and evaluate on the same test set. For reference, there are 12245 images with gender labelled for training, 4007 ones for testing, and 11823 images with age group labelled for training, and 4316 ones for testing.

\begin{figure}[t]
\centering
\includegraphics[width=0.85\linewidth]{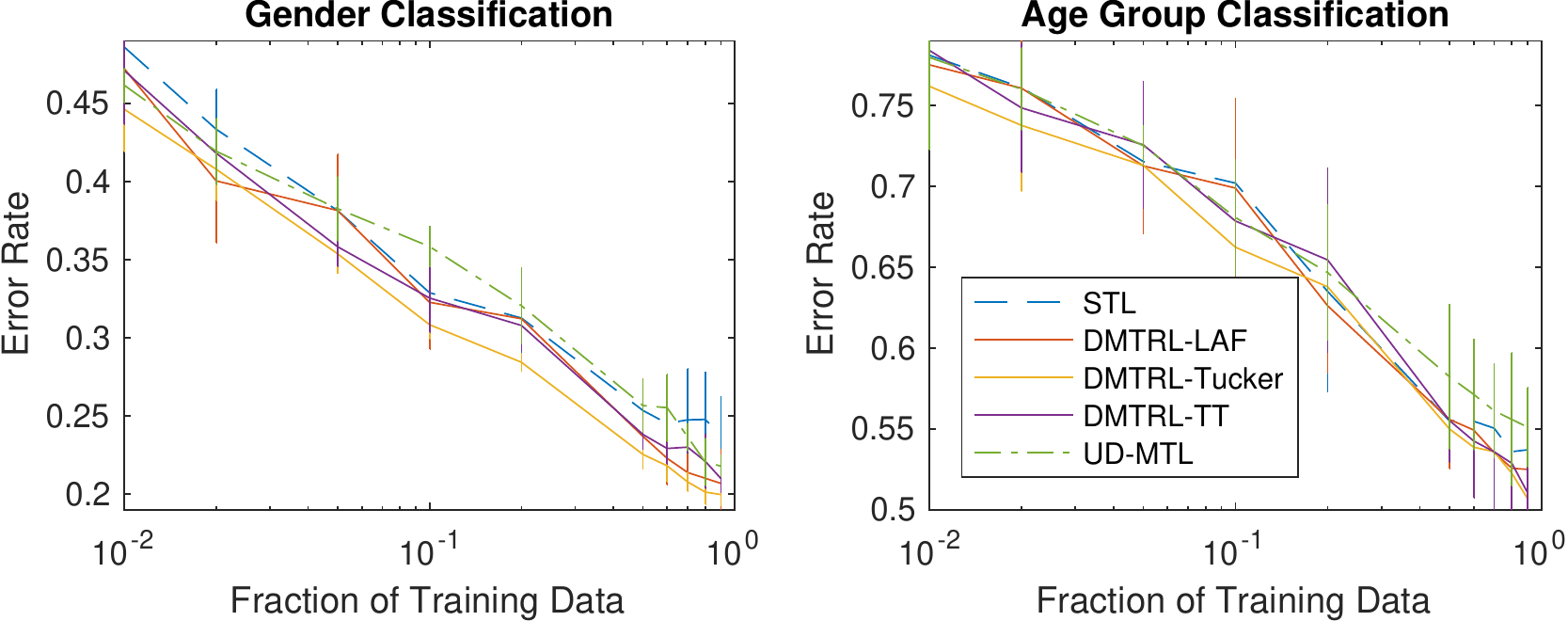}
\caption{Heterogeneous MTL: Age and Gender recognition in AdienceFace dataset.}
\label{fig:ga}
\end{figure}

\textbf{Results}\quad Fig.~\ref{fig:ga} shows the error rate for each task. For the gender recognition task, we find that: (i) User-defined MTL is not consistently better than STL, but (ii) our methods, esp., DMTRL-Tucker, consistently outperform both STL and the best user-defined MTL. For the harder age group classification task, our methods generally improve on STL. However UD-MTL does not consistently improve on STL, and even reduces performance when the training set is bigger. This is the negative transfer phenomenon \citep{Rosenstein05totransfer}, where using a transfer learning algorithm is worse than not using it. This difference in outcomes is attributed to sufficient data eventually providing some effective task-specific representation. Our methods can discover and exploit this, but UD-MTL's hard switch between sharing and not sharing can not represent or exploit such increasing task-specificity of representation.

\subsection{Heterogeneous MTL: Multi-Alphabet Recognition}
\textbf{Dataset, Settings and Baselines}\quad We next consider the task of learning to recognise handwritten letters \emph{in multiple languages} using the Omniglot \citep{Lake1332} dataset. Omniglot contains handwritten characters in 50 different alphabets (e.g., Cyrillic, Korean, Tengwar),  each with its own number of unique characters ($14\sim 55$). In total, there are 1623 unique characters, and each has exactly 20 instances. Here each task corresponds to an alphabet, and the goal is to recognise its characters. MTL has a clear motivation here, as  cross-alphabet knowledge sharing is likely to be useful  as one is unlikely to have extensive training data for a wide variety of less common alphabets.

The images are monochrome of size $105\times 105$. We design a CNN  with $3$ convolutional  and $2$ FC layers. The first conv layer has $8$ filters of size $5\times 5$; the second conv layer has $12$ filters of size $3\times 3$, and the third convolutional layer has $16$ filters of size $3\times 3$. Each convolutional layer is followed by a $2\times 2$ max-pooling. The first FC layer has $64$ neurons, and the second FC layer has size corresponding to the number of unique classes in the alphabet. The activation function is $tanh$.

We use a similar strategy to find the best user-defined MTL model: the CNN has $5$ parametrised layers, of which $4$ layers are potentially shareable. So we tried hard-sharing the first $N$ ($1\leq N \leq 4$) layers. Evaluating these options by $5$-fold cross-validation, the best option turned out to be $N=3$, i.e.,  the first \emph{three} layers are hard shared. For our methods, \emph{all four} shareable layers are softly shared.

Since there is no standard train/test split for this dataset, we use the following setting: We repeatedly pick at random $5,\dots 90\%$ of images per class for training. Note that $5\%$ is the minimum, corresponding to one-shot learning. The remaining data are used for evaluation.

\begin{figure}[t]
\centering
\includegraphics[width=0.85\linewidth]{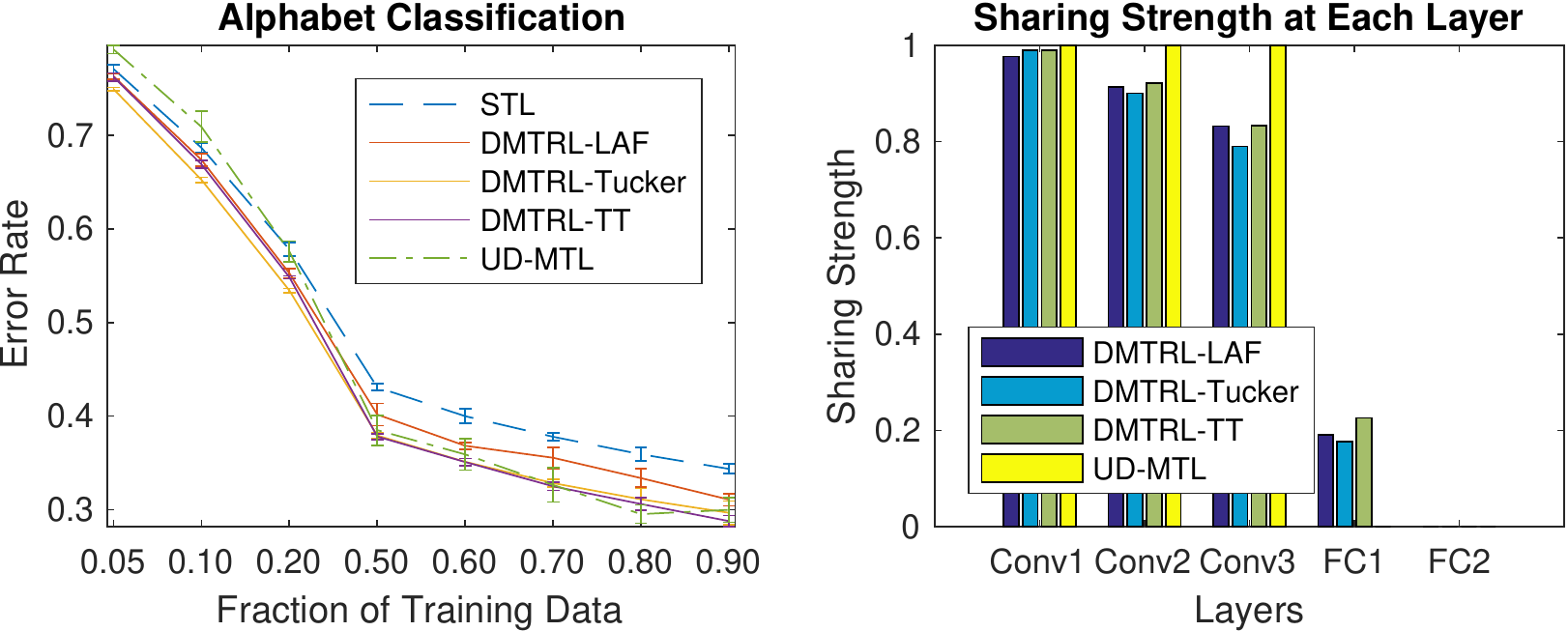}\\
\vspace{0.2cm}
\includegraphics[width=0.85\linewidth]{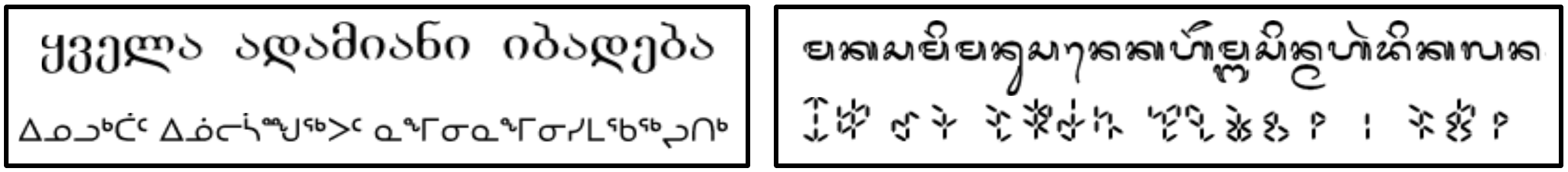}
\caption{Results of multi-task learning of multilingual character recognition (Omniglot dataset).  Below: Illustration of the language pairs estimated to be the most related (left - Georgian Mkhedruli and Inuktitut) and most unrelated (right - Balinese and ULOG) character recognition tasks.}
\label{fig:alphabet}
\end{figure}

\textbf{Results}\quad 
Fig.~\ref{fig:alphabet} reports the average  error rate across all $50$ tasks (alphabets). Our proposed MTL methods surpass the STL baseline in all cases. User-defined MTL does not work well when the training data is very small, but does help when training fraction is larger than $50\%$. 

\textbf{Measuring the Learned Sharing}\quad 
Compared to the conventional user-defined sharing architectures, our method learns how to share from data. We next try to quantify the amount of sharing estimated by our model on the Omniglot data. Returning to the key factorisation $\mathcal{W}=\mathcal{L}S$, we can find that $S$-like matrix appears in all variants of proposed method. It is $S$ in DMTRL-LAF, the transposed ${U^{(N)}}$ in DMTRL-Tucker, and $U^{(N)}$ in DMTRL-TT ($N$ is the last axis of $\mathcal{W}$). $S$ is a $K\times T$ size matrix, where $T$ is the number of tasks, and $K$ is the number of latent tasks \citep{daume2012gomtl} or the dimension of task coding \citep{yang15}. Each column of $S$ is a set of coefficients that produce the final weight matrix/tensor by linear combination. If we put STL and user-defined MTL (for a certain shared layer) in this framework, we see that STL is to \emph{assign} (rather than \emph{learn}) $S$ to be an identity matrix $I_T$. Similarly, user-defined MTL (for a certain shared layer) is to assign $S$ to be a matrix with all zeros but one particular row is all ones, e.g., $S=[\mathbf{1}_{1\times T};\mathbf{0}]$. Between these two extremes, our method learns the sharing structure in $S$. We propose the following equation to measure the learned sharing strength:
\begin{equation}
	\rho = \frac{1}{\binom T2}\sum_{i<j} \Omega(S_{\cdot,i}, S_{\cdot,j}) = \frac{2}{T(T-1)}\sum_{i<j} \Omega(S_{\cdot,i}, S_{\cdot,j})
	\label{eq:rho}
\end{equation}
\hspace{-0.55em} Here $\Omega(a,b)$ is a similarity measure for two vectors $a$ and $b$ and we use cosine similarity. $\rho$ is the average on all combinations of column-wise similarity. So $\rho$ measures how much sharing is encoded by $S$ between $\rho=0$ for STL (nothing to share) and $\rho=1$ for user-defined MTL (completely shared). Since $S$ is a real-valued matrix in our scenario, we normalise it before applying Eq.~\ref{eq:rho}: First we take absolute values, because large either positive or negative value suggests a significant coefficient. Second we normalise each column of $S$ by applying a softmax function, so the sum of every column is $1$. The motivation behind the second step is to make a matched range of our $S$ with $S=I_T$ or $S=[\mathbf{1}_{1\times T};\mathbf{0}]$, as for those two cases, the sum of each column is $1$ and the range is $[0,1]$.

For the Omniglot experiment, we plot the measured sharing amount for training fraction $10\%$. Fig.~\ref{fig:alphabet} reveals that three proposed methods tend to share more for bottom layers (`Conv1', `Conv2', and `Conv3') and share less for top layer (`FC1'). This is qualitatively similar to the best user-defined MTL, where the first three layers are fully  shared ($\rho=1$) and the $4$th layer is completely not shared ($\rho=0$). However, our methods: (i) learn this structure in a purely data-driven way and (ii) benefits from the ability to smoothly interpolate between high and low degrees of sharing as depth increases. As an illustration, Fig.~\ref{fig:alphabet} also shows example text from the most and least similar language pairs as estimated at our multilingual character recogniser's FC1 layer (the result can vary across layers).

\section{Conclusion}

In this paper, we propose a novel framework for end-to-end multi-task representation learning in contemporary deep neural networks. The key idea is to generalise matrix factorisation-based multi-task ideas to tensor factorisation, in order to flexibly share knowledge in fully connected and convolutional DNN layers. Our method provides consistently better performance than single task learning and comparable or better performance than the best results from exhaustive search of user-defined MTL architectures. It reduces the design choices and architectural search space that must be explored in the workflow of Deep MTL architecture design \citep{Caruana1997,ZhangLLT14, Liu15nlp}, relieving researchers of the need to decide how to structure layer sharing/segregation.  Instead sharing structure is determined in a data-driven way on a layer-by-layer basis that moreover allows a smooth interpolation between sharing and not sharing in progressively deeper layers. 

\vspace{0.2cm}\noindent\textbf{Acknowledgements}\quad This work was supported by EPSRC (EP/L023385/1), and the European Union's Horizon 2020 research and innovation program under grant agreement No 640891.

\clearpage
\appendix

\section{Comparison with classic (shallow) MTL methods }
\label{appx2}

We provide a comparison with classic (shallow, matrix-based) MTL methods for the first experiment (MNIST, binary one-vs-rest classification, $1\%$ training data, mean of error rates for 10-fold CV).
A subtlety in making this comparison is what feature should the classic methods use? Conventionally they use a PCA feature (obtained by flattening the image, then dimension reduction by PCA). However for visual recognition tasks, performance is better with deep features -- a key motivation for our focus on deep approaches to MTL. We therefore also compare the classic methods when using a feature extracted from the penultimate layer of the CNN network used in our experiment.

\begin{table}[h]
	\centering
	\begin{tabular}{c|c c}
		\hline
		Model & PCA Feature & CNN Feature\\
		\hline 
		Single Task Learning & 16.89 & 11.52  \\ 

		\cite{Evgeniou2004} & 15.27  & 10.32  \\ 

		\cite{Argyriou2008} & 15.64 & ~~9.56 \\ 

		\cite{daume2012gomtl} & 14.08  & ~~9.41  \\ 
		 
		DMTRL-LAF & - & ~~8.25 \\ 
		 
		DMTRL-Tucker & - & ~~9.24 \\ 
		 
		DMTRL-TT& - & ~~7.31 \\ 
		 
		UD-MTL& - & ~~9.34 \\ 
		\hline 
	\end{tabular} 
	\caption{Comparison with classic MTL methods. MNIST binary classification error rate (\%).}
	\label{tab:other_mtl}
\end{table}

As expected, the classic methods improve on STL, and they perform significantly better with CNN than PCA features. However, our DMTRL methods still outperform the best classic methods, even when they are enhanced by CNN features. This is due to soft (cf hard) sharing of the feature extraction layers and the ability of end-to-end training of both the classifier and feature extractor. Finally, we note that more fundamentally, the classic methods are restricted to binary problems (due to their matrix-based nature) and so, unlike our tensor-based approach, they are unsuitable for multi-class problems like omniglot and age-group classification.

\section{Model capacity and Performance}
\label{appx1}

We list the number of parameters for each model in the first experiment (MNIST, binary one-vs-rest classification) and the performance ($1\%$ training data, mean of error rate for 10-fold CV).

\begin{table}[h]
\centering
\begin{tabular}{c|c c c c}
	\hline 
	Model & Error Rate (\%) & Number of parameters & Ratio  \\ 
	\hline 
	STL & 11.52 & 4351K  & 1.00 \\  
	DMTRL-LAF &  ~~8.25 & 1632K   &  0.38  \\ 

	DMTRL-Tucker &  ~~9.24 & 1740K & 0.40 \\ 
	 
	DMTRL-TT & ~~7.31 & 2187K & 0.50 \\ 
	 
	UD-MTL & ~~9.34 & ~~436K & 0.10 \\ 
	 
	UD-MTL-Large & ~~9.39 & 1644K & 0.38 \\ 
	\hline

\end{tabular} 
	\caption{Comparison of deep models: Error rate and number of parameters.}
	\label{tab:param_num}
\end{table}

The conventional hard-sharing method (UD-MTL) design is to share all layers except the top layer. Its number of parameter is roughly  $10\%$ of the single task learning method (STL), as most parameters are shared across the 10 tasks corresponding to 10 digits. Our soft-sharing methods also significantly reduce the number of parameters compared to STL, but are larger than UD-MTL's hard sharing.

To compare our method to UD-MTL, while controlling for network capacity, we expanded UD-MDL by adding more hidden neurons so its number of parameter is close to our methods (denoted UD-MTL-Large). However UD-MDL performance does not increase. This is evidence that our model's good performance is not simply due to greater capacity than UD-MTL.

\end{document}